\documentclass[10pt, conference, compsocconf]{IEEEtran}
\IEEEoverridecommandlockouts
\usepackage{cite}
\usepackage{amsmath,amssymb,amsfonts}
\usepackage{algorithmic}
\usepackage{graphicx}
\usepackage{textcomp}
\usepackage{xcolor}
\usepackage{comment}
\usepackage{svg}
\def\BibTeX{{\rm B\kern-.05em{\sc i\kern-.025em b}\kern-.08em
    T\kern-.1667em\lower.7ex\hbox{E}\kern-.125emX}}
\begin{document}

\title{Implementation of a modified Nesterov's Accelerated quasi-Newton Method on Tensorflow\\
}

\author{\IEEEauthorblockN{ S. Indrapriyadarsini }
\IEEEauthorblockA{\textit{Graduate School of Integrated Science and Technology, 
} \\
\textit{Shizuoka University, }\\
Hamamatsu, Japan \\
s.indrapriyadarsini.17@shizuoka.ac.jp}
\and
\IEEEauthorblockN{ Shahrzad Mahboubi}
\IEEEauthorblockA{\textit{Graduate School of Electrical and Information Engineering, } \\
\textit{Shonan Institute of Technology,}\\
Fujisawa, Japan \\
18T2012@sit.shonan-it.ac.jp}
\and
\IEEEauthorblockN{Hiroshi Ninomiya}
\IEEEauthorblockA{\textit{Graduate School of Electrical and Information Engineering, } \\
\textit{Shonan Institute of Technology,}\\
Fujisawa, Japan \\
ninomiya@info.shonan-it.ac.jp}
\and
\IEEEauthorblockN{Hideki Asai}
\IEEEauthorblockA{\textit{Research Institute of Electronics,} \\
\textit{Shizuoka University, }\\
Hamamatsu, Japan \\
asai.hideki@shizuoka.ac.jp}
}

\maketitle

\begin{abstract}
Recent studies incorporate Nesterov's accelerated gradient method for the acceleration of gradient based training. The Nesterov's Accelerated Quasi-Newton (NAQ) method has shown to drastically improve the convergence speed compared to the conventional quasi-Newton method. This paper implements NAQ for non-convex optimization on Tensorflow. Two modifications have been proposed to the original NAQ algorithm to ensure global convergence and eliminate linesearch. The performance of the proposed algorithm - mNAQ is evaluated on standard non-convex function approximation benchmark problems and microwave circuit modelling problems. The results show that the improved algorithm converges better and faster compared to first order optimizers such as AdaGrad, RMSProp, Adam, and the second order methods such as the quasi-Newton method.
\end{abstract}


\section{Introduction}
Neural networks have been effective in solving high non-linear problems and thus find many applications in solving real-world problems such as microwave modelling [1][2]. The neural networks can be trained from Electromagnetic (EM) data over a range of geometrical parameters which can be used as models for providing fast solutions of the EM behavior. Such modelling is especially very useful where formulas are not available or original model is computationally too expensive [2-4]. 

Training is the most important step in developing a neural network model. Gradient based algorithms are popularly used in training and can be divided into two categories - first order methods and second or approximated second order methods [1]. There have been several recent advancements in first order optimization methods such as AdaGrad [7], RMSProp [8], Adam [9], etc. These methods are based on stochastic (minibatch) strategies. Stochastic strategies are not suitable for the training of high non-linear functions due to its complex characteristics [10].  Applications such as EM simulation have strong non-linearity and require low training errors. Therefore second-order training methods are more suitable than first-order algorithms. Second order methods though faster in convergence compared to first-order methods, are computationally expensive due to the calculation of the Hessian matrix. In quasi-Newton (QN) methods, the Hessian matrix is computed by iterative approximations. The BFGS algorithm is one of the most popular quasi-Newton methods. Several improvements have been proposed to quasi-Newton methods that result in faster and better convergence [3][5]. These methods obtain stronger local convergence than QN over a long simulation time. Although they improved asymptotic convergence rates, the methods are still often slow in practice. 

There have been several attempts in applying second order quasi-Newton methods for neural network optimization in the famous deep learning framework Tensorflow. TensorFlow enables developers to experiment with novel optimizations and training algorithms. Recently [15] proposed the Nesterov's Accelerated quasi-Newton (NAQ) method which guarantees faster convergence compared to first-order methods and the classical quasi-Newton method. However, on implementing the BFGS quasi-Newton method and NAQ on Tensorflow, we observed frequent terminations caused by failure to determine the stepsize. BFGS and NAQ implement linesearch satisfying conditions such as Wolfe and Armijo respectively for determining the stepsize. Recent studies [11]-[13] show that linesearch methods do not necessarily tend to global convergence. In this paper, we propose two modifications to the original NAQ algorithm that ensures global convergence and eliminate linesearch. 

This paper attempts to study the performance of the proposed algorithm in optimizing neural networks in Tensorflow. We evaluate the performance of the algorithms on  function approximation problems and microwave circuit problems.

\section{Formulation of training and Gradient based training methods}

\subsection{Formulation of Training}

Let ${\bf d}_p$, ${\bf o}_p$ and ${\bf w} \in {\mathbb{R}}^D$ be the {\it p}-th desired ouput, obtained output and weight vectors respectively. The error function $E({\bf w})$ is defined as 
\begin{equation}
E({\bf w}) =\frac{1}{|T_r|}{ \sum_{p \in T_r} E_p ({\bf w})},\;\; E_p ({\bf w})=\frac{1}{2} \|{\bf d}_p-{\bf o}_p\|^2, 
\end{equation}
where ${\it T}_r$ denotes a training data set $\{{\bf x}_p,{\bf d}_p\}, {p \in T_r}$ and $|T_r|$ is the number of training samples. In gradient-based algorithms, the
error function is minimized by the following iterative formula:\vspace {-0.2cm}
\begin{equation}
{\bf w}_{k+1}={\bf w}_k+{\bf v}_{k+1}, 
\end{equation}
where ${\it k}$ is the iteration count and ${\bf v}_{k+1}$ is the  update vector, which is defined for each gradient algorithm. The update vector of the simple steepest gradient algorithm so-called Back-propagation method (BP) [1] is given as 
\begin{equation} 
{\bf v}_{k+1}=-\alpha_k\nabla E({\bf w}_k).
\end{equation}
where $\alpha_k$ is the stepsize and $\nabla E({\bf w}_k)$ is the gradient at ${\bf w}_k$. 

\subsection{Gradient Based Training}
\subsubsection{Classical Momentum} 
The classical momentum method (CM) accelerates BP by accumulating previous vector updates in direction of persistent reduction [6]. The update vector of CM is given by:
\begin{equation}
{\bf v}_{k+1}= \mu {\bf v}_k - \alpha_k \nabla E({\bf w}_k).
\end{equation}
where $\mu \in (0,1)$ denotes the momentum term. \\

\subsubsection{Nesterov's Accelerated Gradient (NAG)}

Nesterov's Accelerated Gradient (NAG) method is a simple modification of CM in which the gradient is computed at ${\bf w}_k + \mu {\bf v}_k$ instead of ${\bf w}_k$ [6]. Thus, the update vector is given by:
\begin{equation}
{\bf v}_{k+1}=\mu_k {\bf v}_k -\alpha_k\nabla E({\bf w}_k + \mu {\bf v}_k).
\end{equation}
where $ \nabla E({\bf w}_k+\mu {\bf v}_k)$ is the gradient at ${\bf w}_k + \mu {\bf v}_k$ and is referred to as Nesterov's accelerated gradient vector.\\

\subsubsection{Adaptive Gradient (AdaGrad)}

Adagrad [7] is an algorithm for gradient-based optimization that adapts the learning rate to the parameters, thus performing smaller updates. The update vector is given as 
\begin{equation}
v_{k+1,i} = - \frac{\alpha}{\sqrt{\mathstrut {\sum_{s=1}^{k}(\nabla E({\bf w}_s)_i)^2}}} \nabla E({\bf w}_k)_i.
\end{equation}
where $v_{k+1,i}$ and $\nabla E({\bf w}_k)_i$ are the {\it i}-th elements of ${\bf v}_{k+1}$ and $\nabla E({\bf w}_k)$, respectively. $\alpha$ is a global stepsize shared by all dimensions. The default value of $\alpha$ is $\alpha=0.01$ [7]. \\

\subsubsection{Root Mean Square propagation (RMSprop)}
The RMSprop optimizer [8] utilizes the magnitude of recent gradients to normalize the gradients. Similar to AdaGrad, the learning rate is adapted for each of the parameters. It computes a moving average over the root mean squared of recent gradients, by which the current gradient is divided. The update vector is given by
\begin{equation}
v_{k+1,i} = - \frac{\alpha}{\sqrt {\mathstrut {\theta_{k,i}+\lambda}}} \nabla E({\bf w}_k)_i,
\end{equation}
where
\begin{equation}
\theta_{k,i} = \gamma \theta_{k-1,i} + (1 - \gamma)(\nabla E({\bf w}_k)_i)^2.
\end{equation}
where $\lambda = 10^{-8}$ and $\theta_{k,i}$ is the parameter of {\it k}-th iteration and $i$-th element. The decay term $\gamma$ and learning rate $\alpha$  are set to 0.9 and 0.001 [8]. \\

\subsubsection{Adam}
Adam is one of the most popular and effective first order methods. It uses exponentially decaying average of past squared gradients and exponentially decaying average of past gradients [9]. The update vector is given as
\begin{equation}
v_{k+1,i}= - \alpha \frac{{\hat m}_{k,i}}{(\sqrt{\mathstrut {\hat \theta}_{i,k}}+ \epsilon)},
\end{equation}
where
\begin{equation}
{\hat m}_{k,i}=\frac{m_{i,k}}{(1- \beta_1^k)},
\end{equation}
\begin{equation}
{\hat \theta}_{k,i} = \frac{\theta_{k,i}}{(1-\beta_2^k)},
\end{equation}
here, $m_{k,i}$ and $\theta_{k,i}$ given by:
\begin{equation}
m_{k,i}=\beta_1 m_{k-1,i}+ (1 - \beta_1) \nabla E({\bf w}_k)_i,
\end{equation}
\begin{equation}
\theta_{k,i} = \beta_2 \theta_{k-1,i}+ (1 - \beta_2) )\nabla E({\bf w}_k)_i)^2.
\end{equation} 
where $\epsilon = 10^{-8}$ and $\beta_1^k$ and $\beta_2^k$ denote the {\it k}-th power of $\beta_1$ and $\beta_2$, respectively. The default value of the global stepsize $\alpha$  is $0.001$ [13]. $m_{k,i}$ and $\theta_{k,i}$ are {\it i}-th elements of the gradient and the squared gradient, respectively. The hyper-parameters $0\leq \beta_1, \beta_2 <1$ control the exponential decay rates of these running averages. The running average themselves are estimates of the first (the mean) moment and the second raw (the uncentered variance) moment of the gradient. $\beta_1$ and $\beta_2$ are chosen to be 0.9 and 0.999, respectively [9]. All operations on vectors are element-wise.

AdaGrad, RMSprop, Adam are based on stochastic strategies and hence not suitable for highly non-linear problems [4][10]. Therefore, we focus on the methods using the curvature information and the full batch strategy in this paper.\\

\subsubsection{Modified quasi-Newton Method}

quasi-Newton methods utilize the gradient of the objective function to result in superlinear quadratic convergence. The Broyden-Fletcher-Goldfarb-Shanon (BFGS) algorithm is one of the most popular quasi-Newton methods for unconstrained optimization [10]-[15]. The update vector of the quasi-Newton (QN) method is given as
\begin{equation}
{\bf v}_{k+1} = \alpha_k {\bf g}_k,
\end{equation}
\begin{equation}
{\bf g}_k = -{\bf H}_k \nabla E({\bf w}_k).
\end{equation}
The hessian matrix ${\bf H}_k$ is symmetric positive definite and is iteratively approximated by the following BFGS formula [13].
\begin{equation}
{\bf H}_{k+1}= ( {\bf I}- \rho_k {\bf s}_k {\bf y}_k^{\rm T}){\bf H}_k({\bf I}- \rho_k {\bf y}_k {\bf s}_k^{\rm T})+ \rho_k {\bf s}_k {\bf s}_k^{\rm T},
\end{equation}
where ${\bf I}$ denotes identity matrix and,
\begin{equation}
{\bf s}_k = {\bf w}_{k+1} - {\bf w}_k,
\end{equation}
\begin{equation}
{\bf y}_k = \nabla E ( {\bf w}_{k+1} ) - \nabla E ({\bf w}_k), 
\end{equation}
\begin{equation}
\rho_k = \frac{1}{{\bf y}_k^{\rm T} {\bf s}_k}.
\end{equation}

The above BFGS method is implemented in scipy and used in Tensorflow through the ScipyOptimizerInterface class. On simulation, we observed that the BFGS implementation did not terminate on convergence but terminated with precision loss error. The error traces back to not being able to determine a suitable stepsize. Hence this implementation is not stable and cannot be used to obtain convergence for all problems.

Many studies have been proposed on the global convergence of quasi-Newton methods. Li and Fukushima [12] suggested a modified BFGS (mBFGS) method by incorporating an additional term $\xi_k {\bf s}_k$ in calculating vector ${\bf y}_k$. This modification confirms global and superlinear convergence without the convexity assumption on function {\it f}. The vector ${\bf y}_k$ in the modified method is given as
\begin{equation}
{\bf y}_k = \nabla E ( {\bf w}_{k+1} ) - \nabla E ({\bf w}_k)+ \xi_k {\bf s}_k = \epsilon_k + \xi_k {\bf s}_k, 
\end{equation}
and $\xi_k$ is defined as
\begin{equation}
\xi_k = \omega \parallel \nabla E({\bf w}_k) \parallel + max\{ {- \epsilon_k^{\rm T} {\bf s}_k}/{ \parallel {\bf s}_k \parallel^2},0\},
\end{equation} 
\begin{equation}
\begin{cases}
\begin{split}
\omega =2~~~~~~~if \parallel \nabla E({\bf w}_k) \parallel^2 >10^{-2},\\
\omega =100~~~~if \parallel \nabla E({\bf w}_k) \parallel^2 <10^{-2}.
\end{split}
\end{cases}
\end{equation}
Further, [14] suggest a method to eliminate line search and determine the stepsize $\alpha_k$ using an explicit formula given by:
\begin{equation}
\alpha_k = -\frac{\delta \nabla E({\bf w}_k)^{\rm T}{\bf g}_k}{||{\bf g}_k||_{Q_k}^2},
\end{equation}
where
\begin{equation}
||{\bf g}_k||_{Q_k}^2 = \sqrt[]{\mathstrut {\bf g}_k^{\rm T} Q_k {\bf g}_k}.
\end{equation}
$Q_k$ is determined by $Q_k={\it L}{\bf I}$ where ${\it L}$ is the Lipschtz constant of the gradient. ${\it L}$ is chosen to be  ${\it L}= 100||{\bf y}_k||/||{\bf s_k}||$. 

The modified quasi-Newton (mBFGS) algorithm is shown in Algorithm 1. We implemented this algorithm in Tensorflow and the simulation results show convergence.

\begin{figure}[htb]
\fontsize{9.5}{13}\selectfont
\hrule height 0.01mm depth 0.1mm width 88mm
\noindent
\\ Algorithm 1:  modified quasi-Newton Method (mBFGS)\\
1. $k = 1;$\\
2. Initialize ${\bf w}_1 = rand[-0.5,0.5]$(uniform random numbers) and ${\bf H}_1$to identity matrix ;\\
3. Calculate $\nabla E({\bf w}_1)$;\\
4. {\bf While}$~ (||E({\bf w}_k|| >{\bf \varepsilon} \;\;{\rm and}\;\; k < k_{max})$\\
\indent\;\;\;\;\;\;\;\;(a)\;Calculate ${\bf g}_k=-{\bf H}_k \nabla E({\bf w}_k)$;\\
\indent\;\;\;\;\;\;\;\;(b)\;{\it if} $E({\bf w}_k +{\bf g}_k) \leq E({\bf w}_k) + \sigma \alpha_k \nabla E({\bf w}_k)^{\rm T} {\bf g}_k$,\\
\indent\;\;\;\;\;\;\;\;\;\;\;\;\;{\it then} $\alpha_k=1$, {\it else} $\alpha_k$ is computed by (23);\\
\indent\;\;\;\;\;\;\;\;(c)\;Update ${\bf w}_{k+1}={\bf w}_k +\alpha_k {\bf g}_k$;\\
\indent\;\;\;\;\;\;\;\;(d)\;Calculate $\nabla E({\bf w}_{k+1})$;\\
\indent\;\;\;\;\;\;\;\;(e)\;Update ${\bf H}_{k+1}$ using (16);\\
\indent\;\;\;\;\;\;\;\;(f)\;$k =k+1$;\\
5. {\bf return}\;${\bf w}_k$;
\hrule height 0.001mm depth 0.1mm width 88mm
\end{figure}

\section{Proposed Algorithm - Modified Nesterov's Accelerated Quasi-Newton method}
\subsection{Nesterov's Accelerated quasi-Newton (NAQ) method}
Several modifications have been proposed to quasi-Newton to obtain stronger convergence. The Nesterov's Accelerated quasi-Newton method [15] gives faster convergence compared to the standard quasi-Newton methods. NAQ obtains faster convergence by quadratic approximation at ${\bf w}_k+\mu {\bf v}_k$ and by incorporating the Nesterov's accelerated gradient $\nabla E({\bf w}_k+\mu {\bf v}_k)$ The derivation of NAQ is briefly introduced as follows:

Let $\Delta {\bf w}$ be the vector $\Delta {\bf w}={\bf w}-\left({\bf w}_k+\mu_k{\bf v}_k \right)$, the quadratic approximation of (1) around ${\bf w}_k+\mu_k{\bf v}_k$ is derived as,
\begin{equation}
\begin{split}
E\left({\bf w}\right) \simeq E\left({\bf w}_k+\mu_k{\bf v}_k \right)+ \nabla E\left({\bf w}_k+\mu_k{\bf v}_k \right)^{\rm T} \Delta{\bf w}\\+\frac{1}{2} \Delta {\bf w}^{\rm T} \nabla^2 E\left({\bf w}_k+\mu_k{\bf v}_k \right) \Delta {\bf w}.
\end{split}
\end{equation}
\noindent
The minimizer of this quadratic function is explicitly given by $\Delta {\bf w}=-\nabla^2 E\left({\bf w}_k+\mu_k{\bf v}_k \right)^{-1} \nabla E\left({\bf w}_k+\mu_k{\bf v}_k \right)$. Therefore the new iterate is defined as
\begin{equation}
\begin{split}
{\bf w}_{k+1}=\left({\bf w}_k+\mu_k{\bf v}_k \right)\;\;\;\;\;\;\;\;\;\;\;\;\;\;\;\;\;\;\;\;\;\;\;\;\;\;\;\;\;\;\;\;\;\;\;\;\;\;\;\;\;\\\;\;\;\;\;\;\;\;\;\;\;\;\;\;\;\;
-\nabla^2 E\left({\bf w}_k+\mu_k{\bf v}_k \right)^{-1} \nabla E\left({\bf w}_k+\mu_k{\bf v}_k \right).
\end{split}
\end{equation}
\noindent
This iteration is considered as Newton method with the momentum term $\mu{\bf v}_k$. The inverse of Hessian $\nabla^2 E({\bf w}_k + \mu_k {\bf v}_k)$ is approximated by the matrix ${\bf H}_{k+1}$ using the update equation
\begin{equation}
{\bf {\hat H}}_{k+1}= ( {\bf I}- {\hat \rho}_k {\bf p}_k {\bf q}_k^{\rm T}){\bf {\hat H}}_k({\bf I}- {\hat \rho}_k {\bf q}_k {\bf p}_k^{\rm T})+ {\hat \rho}_k {\bf p}_k {\bf p}_k^{\rm T},
\end{equation}
where 
\begin{equation}
{\bf p}_k = {\bf w}_{k+1} - ({\bf w}_k+ \mu_k {\bf v}_k),
\end{equation}
\begin{equation}
{\bf q}_k = \nabla E ( {\bf w}_{k+1} ) - \nabla E ({\bf w}_k+ \mu_k {\bf v}_k),
\end{equation}
\begin{equation}
{\hat \rho}_k = \frac{1}{{\bf q}_k^{\rm T} {\bf p}_k}.
\end{equation}
Equation (27) is derived from the secant condition given below
\begin{equation}
{\bf q}_k=({\bf H}_{k+1})^{-1}{\bf p}_k.
\end{equation}
and the rank-2 updating formula [15]. Note that it is proved that the matrix ${\bf H}_{k+1}$ updated by (27) is a positive definite symmetric matrix given as ${\bf H}_k$ is one. Therefore, the update vector of NAQ can be written as:
\begin{equation}
{\bf v}_{k+1} = \mu_k {\bf v}_k - \alpha_k {\bf {\hat g}}_k,
\end{equation} 
\begin{equation}
 {\bf {\hat g}}_k=-{\bf {\hat H}}_k \nabla E({\bf w}_k+\mu_k {\bf v}_k).
\end{equation} 

We first implemented the NAQ algorithm on Tensorflow using the scipy-BFGS as base class. Similar to BFGS it was noticed that the algorithm implemented on Tensorflow often terminated with precision loss error which again traced back to failure to determine a suitable stepsize. To stabilize the NAQ implementation, we propose two modifications to the original NAQ algorithm. First incorporating the $\xi_k{\bf s}_k$ term and second eliminating line search to ensure global convergence and subsequently reduce the number of function evaluations.

\subsection{Global Convergence}

The line search methods satisfying Armijo's condition or Wolfe conditions does not necessarily ensure global convergence[13]. In the proposed algorithm, we incorporate an additional $\xi_k {\bf s}_k$ term in (29) of the original NAQ algorithm to ensure global convergence[14]. Thus, the vector ${\bf q}_k$ in the modified method is given as
\begin{equation}
{\bf q}_k = \nabla E ( {\bf w}_{k+1} ) - \nabla E ({\bf w}_k+ \mu {\bf v}_k)+ {\hat \xi}_k {\bf p}_k = \epsilon_k + {\hat \xi}_k {\bf p}_k.
\end{equation}
where ${\hat \xi}_k$ is defined as
\begin{equation}
{\hat \xi}_k = \omega \parallel \nabla E({\bf w}_k+ \mu {\bf v}_k) \parallel + max\{ {- \epsilon_k^{\rm T} {\bf p}_k}/{ \parallel {\bf p}_k \parallel^2},0\},
\end{equation} 
\begin{equation}
\begin{cases}
\begin{split}
\omega =2~~~~~~~if \parallel \nabla E({\bf w}_k+ \mu {\bf v}_k) \parallel^2 >10^{-2},\\
\omega =100~~~~if \parallel \nabla E({\bf w}_k+ \mu {\bf v}_k) \parallel^2 <10^{-2}.
\end{split}
\end{cases}
\end{equation}

\subsection{Elimination of Line Search}
The original NAQ algorithm applies backtracking line search that satisfies the Armijo's condition as given in equation (37). 
\begin{equation}
E({\bf w}_k + \mu {\bf v}_k + \alpha_k {\bf {\hat g}}_k) \leq E({\bf w}_k + \mu {\bf v}_k) + \eta \alpha_k \nabla E({\bf w}_k + \mu {\bf v}_k)^{\rm T} {\bf {\hat g}}_k,
\end{equation}
\noindent
where $0 < \eta < 1$ and default value is $\eta =0.001 $.

[14] show that line search satisfying Armijo or Wolfe conditions does not necessarily ensure global convergence Furthermore, linesearch for determining the stepsize  $\alpha_k$ involves an additional computation of $E({\bf w}_k+\mu {\bf v}_k+ \alpha_k  {\bf {\hat g}}_k)$ at each iteration until a suitable stepsize is determined, thereby further increasing the number of function evaluations. Also, it not feasible to fix $\alpha_k$ to a constant value throughout all iterations, as this does not ensure convergence. Hence, we use an explicit formula (38) for determining the stepsize $\alpha_k$. Thus line search is eliminated and the stepsize $\alpha_k$ is determined using the formula 
\begin{equation}
\alpha_k = -\frac{\delta \nabla E({\bf w}_k+\mu {\bf v}_k)^{\rm T}{\bf {\hat g}}_k}{||{\bf {\hat g}}_k||_{Q_k}^2},
\end{equation}
where
\begin{equation}
||{\bf {\hat g}}_k||_{Q_k} = \sqrt[]{\mathstrut {\bf {\hat g}}_k^{\rm T} Q_k {\bf {\hat g}}_k}.
\end{equation}
$Q_k$ is determined by $Q_k={\it L}{\bf I} $ where ${\it L}$ is the Lipschtz constant of the gradient. ${\it L}$ is chosen to be  ${\it L}= 100||{\bf q}_k||/||{\bf p}_k||$.
 
The proposed  mNAQ shown in Algortithm 2 is implemented on Tensorflow.

\begin{figure}[htb]
\fontsize{10}{13}\selectfont

\hrule height 0.01mm depth 0.1mm width 88mm
\noindent
\\ Algorithm 2:  Proposed Algorithm (mNAQ) \\
1. $k = 1;$\\
2. Initialize ${\bf w}_1 = rand[-0.5,0.5]$ (uniform random numbers), ${\bf v}_1 = 0$ and ${\bf H}_1$to identity matrix ;\\
3. {\bf While}$~(||E({\bf w}_k)|| >{\bf \varepsilon} \;\;{\rm and}\;\; k < k_{max})$\\
\indent\;\;\;\;\;\;\;\;(a)\;Calculate $\nabla E({\bf w}_k+\mu {\bf v}_k)$;\\
\indent\;\;\;\;\;\;\;\;(b)\;Calculate ${\bf {\hat g}}_k=-{\bf {\hat H}}_k \nabla E({\bf w}_k+\mu {\bf v}_k)$;\\
\indent\;\;\;\;\;\;\;\;(c)\;{\it if} $E({\bf w}_k + \mu {\bf v}_k+ {\bf {\hat g}}_k) \leq\\
\indent\;\;\;\;\;\;\;\;\;\;\;\;\;\;\;\; E({\bf w}_k+\mu {\bf v}_k) + \sigma \alpha_k \nabla E({\bf w}_k+\mu {\bf v}_k)^{\rm T} {\bf {\hat g}}_k$,\\
\indent\;\;\;\;\;\;\;\;\;\;\;\;\;{\it then} $\alpha_k=1$, {\it else} $\alpha_k$ is computed by (38);\\
\indent\;\;\;\;\;\;\;\;(d)\;Update ${\bf v}_{k+1}=\mu {\bf v}_k +\alpha_k {\bf {\hat g}}_k$;\\
\indent\;\;\;\;\;\;\;\;(e)\;Update ${\bf w}_{k+1}={\bf w}_k +{\bf v}_{k+1}$;\\
\indent\;\;\;\;\;\;\;\;(f)\;Calculate $\nabla E({\bf w}_{k+1})$;\\
\indent\;\;\;\;\;\;\;\;(g)\;Update ${\bf \hat H}_{k+1}$ using (27);\\
\indent\;\;\;\;\;\;\;\;(h)\;$k =k+1$;\\
5. {\bf return}\;${\bf w}_k$;

\hrule height 0.001mm depth 0.1mm width 88mm
\end{figure}

\section{Simulation Results}

The performance of the proposed algorithm is evaluated on two non-convex function approximation problems and two microwave circuit modelling problems. The simulations are performed in Tensorflow. The implementation of mBFGS, NAQ and mNAQ are built upon scipy-BFGS and the performance is compared against AdaGrad [9], RMSProp [10], Adam [12] using Tensorflow's built-in optimizers. The neural network used is a simple two-layer feedforward network with sigmoid activation function. For each example, 15 independent runs are performed. Each trained neural network is estimated by average, best and worst of $E({\bf w}_k)$, average computation time $T$ in seconds and average number of iterations. For each example discussed below, the maximum number of iterations $k_{max}$  is chosen to be $k_{max}=100000$ and the terminate condition is set to $\epsilon = 1.0\times 10^{-6}$. The parameter $\sigma$ is chosen to be  $\sigma= 10^{-3}$. Each element of the input and desired outputs of the training and test data are normalized. In the range$ [-1,1]$. The hyper-parameters of AdaGrad, RMSProp and Adam are set to their default values. The simulation results with a momentum term $\mu$ of 0.8, 0.85, 0.9 and 0.95 are further discussed below. 
  
\begin{table*}[htb]
\vspace{-0.3cm}
\begin{center}
\caption{Summary of simulation results of example 1.}
\label{tab:style}
\fontsize{9}{11}\selectfont
\begin{tabular}{ccccccc}
\hline 
Algorithm & $\mu$ & $E({\bf w})$($\times 10^{-3})$&Time & Iteration&$E_{test}({\bf w})$($\times 10^{-3})$  \\
&&Ave/Best/Worst& (s)& count&Ave/Best/Worst\\
\hline
AdaGrad & - & 59.8 / 58.6 / 60.2 & 40 & 100,000& 59.03 / 57.69 / 59.48 \\
\hline
RMSprop & - & 3.34 / 0.564 / 7.89 & 41 & 100,000 & 3.35 / 0.409 / 8.16\\
\hline
Adam & - & 4.15 / 0.324 / 14.3 & 42 & 100,000 & 4.14 / 0.359 / 14.53\\
\hline
BFGS & - & 15.14 / 0.650 / 31.80 & 4.9 & 3,204 & 15.14 / 0.650 / 30.66\\
\hline
mBFGS & - &5.24 / 0.194 / 17.8 & 58& 31,370 & 5.26 / 0.233 / 17.80\\
\hline
& 0.8 &1.94 / 0.307 / 6.33 & 23 & 9,006& 1.94 / 0.307 / 6.33\\
\cline{2-6}
mNAQ & 0.85 & 0.974 / 0.307 / 5.00 & 19 & 7,549& 0.980 / 0.315 / 5.00 \\
\cline{2-6}
 & 0.9 &1.53 / 0.194 / 13.8 &15 & 5,931& 1.53 / 0.194 / 13.80 \\
\cline{2-6}
 & 0.95 & 1.30 / 0.195 / 6.31 &11 & 4,461& 1.30 / 0.233 / 6.31\\
\hline
\end{tabular}
\end{center}
\vspace{-0.3cm}
\end{table*}
\subsection{Benchmark Problems}
\subsection*{\indent$<{\it Example~1}>$}
The function approximation problem under consideration is given as 
\begin{equation}
f(a, x, b)=1+(x+2x^2 ){\rm sin}(-ax^2+b ),   \hspace{1cm}    |x| \leq 4.   
\end{equation}
Consider the case where $a= -1$ and $b=0$. Thus, the function reduces to a single input function in $x$ given by $f (x) =1+(x+2x^2){\rm sin}(-x^2)$. The training samples are generated with an interval of 0.02 while the test samples are generated by random sampling in the range $x\in [-4,4)$. The training and test set consists of 400 and 10000 samples respectively. The number of hidden neurons used is 7.  Thus, the neural network structure is given as 1-7-1. The summary of the results is presented in Table I. NAQ failed to determine a suitable step size and hence terminated much earlier without converging. Thus, the corresponding results are omitted from the table. The results indicate that the second order methods, mBFGS and mNAQ converge faster with smaller errors compared to the first order algorithms. On comparing the second order methods, mBFGS results are comparable with mNAQ. However, it is 12 times slower and takes almost 9.9 times more number of epochs to converge (Fig. 1) The proposed mNAQ algorithm results in 5-7 times smaller error rates compared to BFGS. On comparing the results of mNAQ with different values of momentum term, $\mu= 0.95$ is the fastest with least number of average epochs while 0.85 has the least average training error. Fig. 2 illustrates the output of the function under consideration versus the output of the neural network trained using mNAQ with a momentum of $\mu = 0.85$. The neural network output is in close approximation with the original function output. Thus, we can conclude that the proposed algorithm can be effectively used to model function approximation problems.

\begin{figure}[htb]
\vspace{-0.5cm}
\begin{center}
\includegraphics[width=6.5cm]{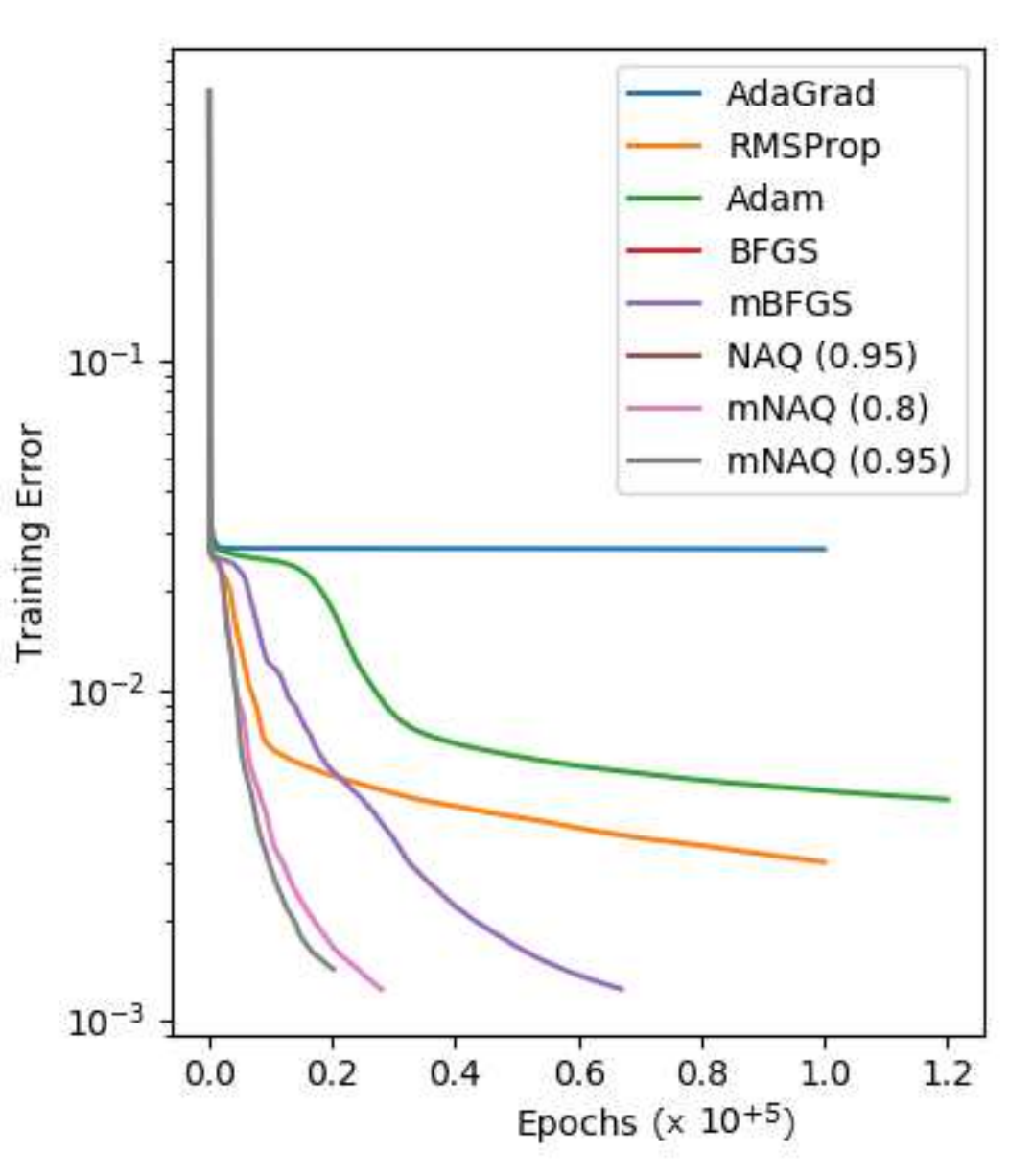}
\end{center}
\vspace{-0.5cm}
\caption{The average training errors $vs$ iteration count for example 1.}
\label{fig:scanner}
\vspace{-0.5cm}
\end{figure}

\begin{figure}[htb]
\begin{center}
\includegraphics[width=7cm]{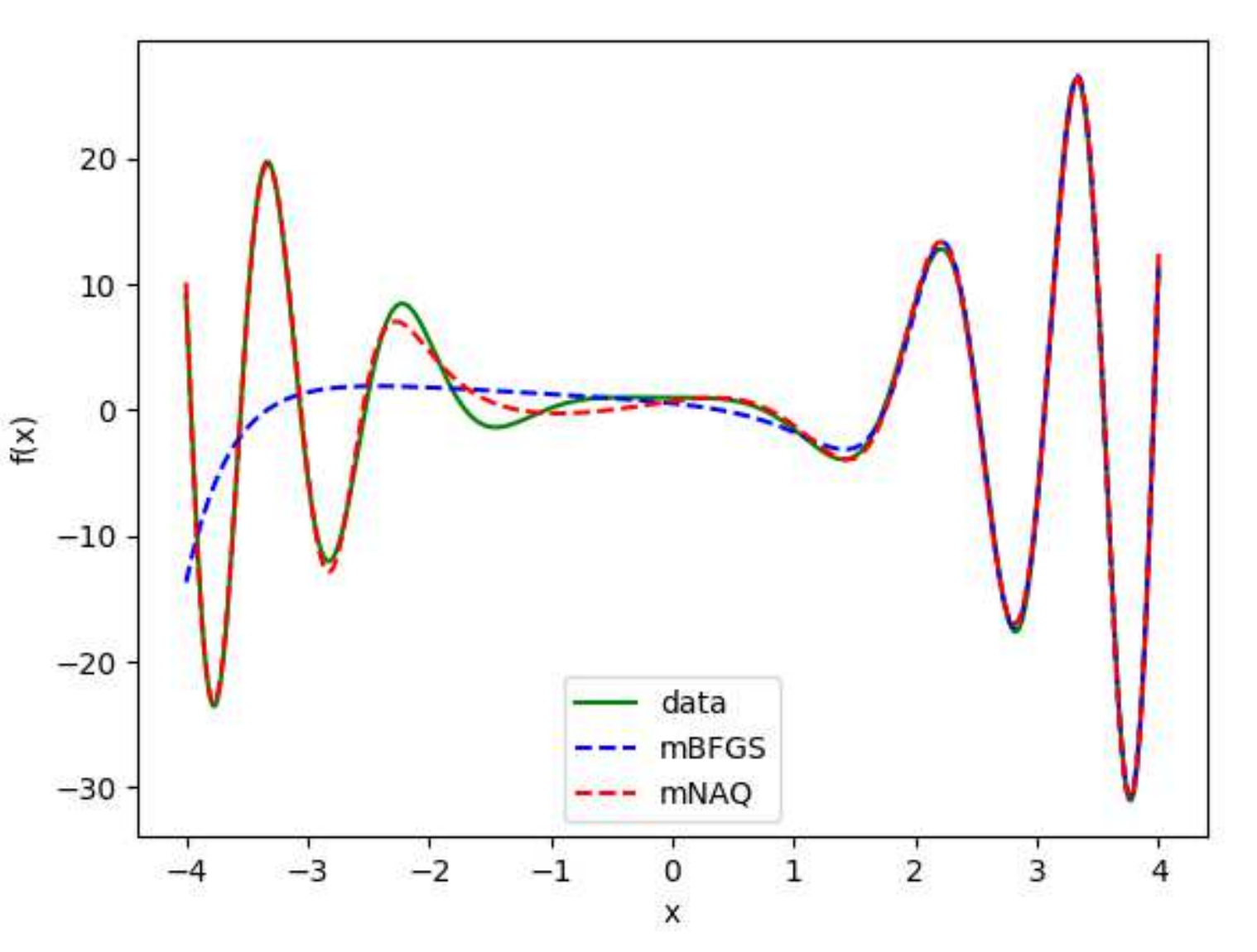}
\end{center}
\vspace{-0.5cm}
\caption{The comparison of network models of mBFGS and mNAQ $vs$ original test data of example 1.}
\label{fig:scanner}
\vspace{-0.5cm}
\end{figure}

\subsection*{\indent$<{\it Example~2}>$}

We further verify the robustness of the algorithm by extending the problem in {\it Example 1}. Consider equation 40 with two-input variables $x\in [-4,4)$ and  $a\in [-4,4)$. The output of the function is given by $f(a, x)=1+(x+2x^2){\rm sin}(ax^2 )$. The training and test dataset comprises of 1680 and 3380 samples respectively.  The number of hidden neurons used is 45. The network structure is 2-45-1 with 181 parameters. Table II shows the results of the simulation. In this example, both BFGS and NAQ failed to determine a suitable step size and hence terminated early without convergence. This further confirms that line search does not tend to converge for all cases and the modified algorithms can resolve this situation. The results corresponding BFGS and NAQ are thus omitted from the table. From the results, we can observe that though Tensorflow's built-in implementation of the first order methods ensures faster runtime, the training and test errors are large compared to the second order methods. The second order methods result in much smaller training and test errors. Further comparing the results of second order methods, training and test errors of mBFGS method is comparable with the proposed algorithm, but mNAQ converges much faster. Fig.3 shows the average training error versus the number of iterations. mNAQ with $\mu =0.85$ has the least average training error.

\begin{figure}[htb]
\vspace{-0.2cm}
\begin{center}
\includegraphics[width=7cm]{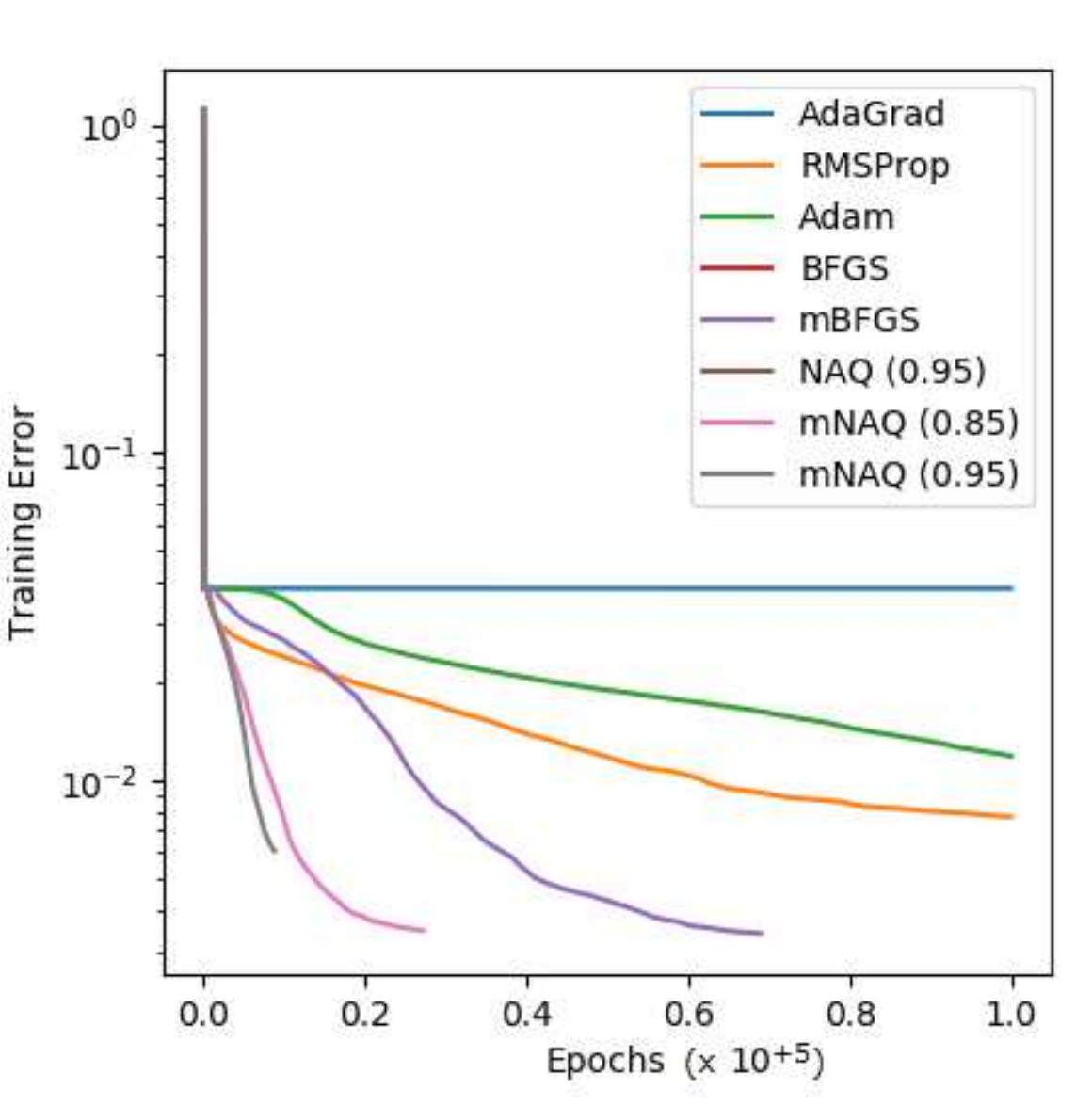}
\end{center}
\vspace{-0.5cm}
\caption{The average training errors $vs$ iteration count for example 2.}
\label{fig:scanner}
\vspace{-0.3cm}
\end{figure}

\begin{table*}[htb]
\begin{center}
\caption{Summary of simulation results of example 2.}
\label{tab:style}
\fontsize{9}{11}\selectfont
\begin{tabular}{ccccccc}
\hline 
Algorithm & $\mu$ & $E({\bf w})$($\times 10^{-3})$&Time & Iteration&$E_{test}({\bf w})$($\times 10^{-3})$  \\
&&Ave/Best/Worst& (s)& count&Ave/Best/Worst\\
\hline
AdaGrad & - & 38.44 / 38.43 / 38.45 & 106 & 100,000& 38.44 / 38.43 / 38.45  \\
\hline
RMSprop & - & 7.71 / 5.33 / 13.95 & 106 & 100,000& 7.75 / 5.40 / 14.00  \\
\hline
Adam & - & 11.82 / 8.03 / 15.71 & 108 & 100,000& 11.82 / 8.03 / 15.71  \\
\hline
mBFGS & - &2.71 / 0.888 / 5.13 & 467 & 84,858& 2.70 / 0.900 / 5.10  \\
\hline
& 0.8 & 2.92 / 0.544 / 4.70 & 375 & 51,135& 2.92 / 0.544 / 4.70 \\
\cline{2-6}
mNAQ & 0.85 & 2.85 / 0.491 / 4.68 & 373 & 50,923& 2.85 / 0.491 / 4.68  \\
\cline{2-6}
 & 0.9 &3.03 / 0.657 / 5.86 &290 & 39,536& 3.03 / 0.657 / 5.86   \\
\cline{2-6}
 & 0.95 & 4.38 / 2.00 / 6.19 &157 & 21,295& 4.38 / 2.00 / 6.19 \\
\hline
\end{tabular}
\end{center}
\end{table*}

\subsection{Modelling of Microwave Circuit Problems}

Neural networks can find application in solving several real-world problems such as microwave circuit modelling. Microwave circuit modelling problems are highly non-linear with many irregularly aligned poles in the S parameter. Thus, modelling of these poles with very small errors is important. In this section, we evaluate the robustness and effectiveness of the proposed algorithm in the modelling of microwave circuits.

\subsection*{\indent$<{\it Example~1: Modelling~of~a~waveguide~filter}>$}

In this example we develop a neural network model of a rectangular waveguide filter (WGF) [16]. The inputs to the neural network are the post distance d and frequency f. The structure of the rectangular waveguide under consideration is shown in Fig 4.  For the training data, length $d = \{3.88, 3.92, 3.96, 4.00$ and $4.04 mm\}$ and for test data length $d = \{3.90, 3.94, 3.98$ and $4.02 mm\}$. Frequency f ranges between 35 to 39 GHz each containing 251 frequency points. The number of training and test samples are 1255 and 1004 respectively.  The outputs are the magnitudes of the S-parameters $|S_{11}|$ and $|S_{21}|$. Fig. 5 shows the training data of the waveguide filter. The number of hidden neurons used is 8. The simulation results are shown in Table III. BFGS and NAQ failed to determine a suitable step size and hence not tabulated. The results of the Adam and RMSProp are comparable with the mBFGS and mNAQ methods. However, the mNAQ converges faster compared to the first order and mBFGS methods. Fig. 6 shows the average training error over epochs. Among the second order methods, mNAQ with $\mu=0.85$ performs the best the least average training error.
\begin{figure}[htb]
\vspace{-0.2cm}
\begin{center}
\includegraphics[width=8cm]{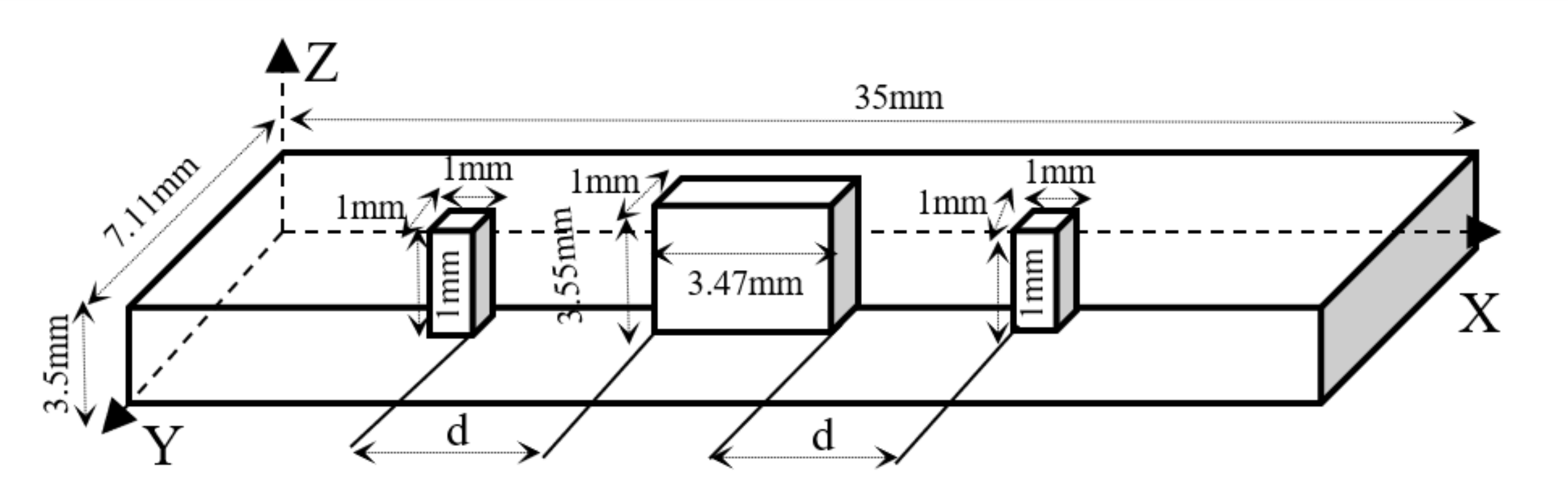}
\end{center}
\vspace{-0.5cm}
\caption{Layout of rectangular waveguide filter (WGF).}
\label{fig:scanner}
\end{figure}

\begin{figure}[htb]
\begin{center}
\includegraphics[width=8cm]{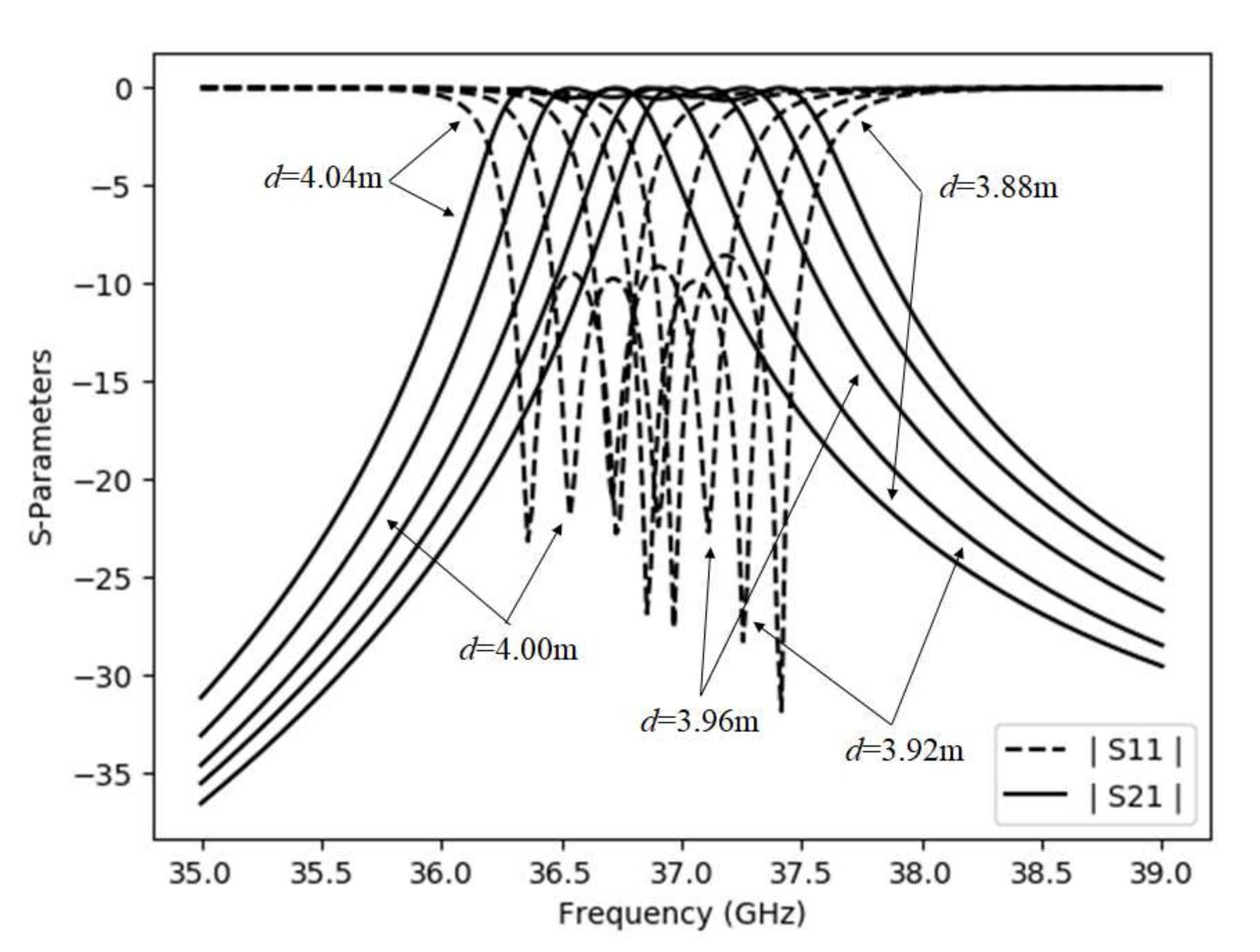}
\end{center}
\vspace{-0.5cm}
\caption{Training data set of WGF.}
\label{fig:scanner}
\vspace{-0.3cm}
\end{figure}

\begin{table*}[htb]
\begin{center}
\caption{Summary of simulation results of rectangular waveguide filter (WGF).}
\label{tab:style}
\fontsize{9}{11}\selectfont
\begin{tabular}{ccccccc}
\hline 
Algorithm & $\mu$ & $E({\bf w})$($\times 10^{-3})$&Time & Iteration&$E_{test}({\bf w})$($\times 10^{-3})$  \\
&&Ave/Best/Worst& (s)& count&Ave/Best/Worst\\
\hline
AdaGrad & - & 73.70 / 14.90 / 105.5 & 69 & 100,000 & 75.51 / 16.6 / 107.1 \\
\hline
RMSprop & - & 0.978 / 0.886 / 1.46 & 70 & 100,000& 1.47 / 1.34 / 2.06   \\
\hline
Adam & - & 1.14 / 0.874 / 3.20 & 70 & 100,000& 1.61 / 1.03 / 1.85  \\
\hline
mBFGS & - &1.03 / 0.856 / 1.38 &81& 21,236& 1.57 / 1.34 / 1.89  \\
\hline
& 0.8 &1.07 / 0.867 / 1.53 &54 & 10,236& 1.59 / 1.40 / 2.13 \\
\cline{2-6}
mNAQ & 0.85 & 0.980 / 0.861  / 1.27 & 45 & 8,442& 1.49 / 1.36 / 1.78  \\
\cline{2-6}
 & 0.9 &1.29 / 0.858 / 4.48 &53 & 9,982&  2.16/ 1.36 / 2.46  \\
\cline{2-6}
 & 0.95 & 1.14 / 0.837 / 1.70 &53 & 10,070& 1.68 / 1.36 / 2.36 \\
\hline
\end{tabular}
\end{center}
\end{table*}

\begin{figure}[htb]
\vspace{-0.4cm}
\begin{center}
\includegraphics[width=7cm]{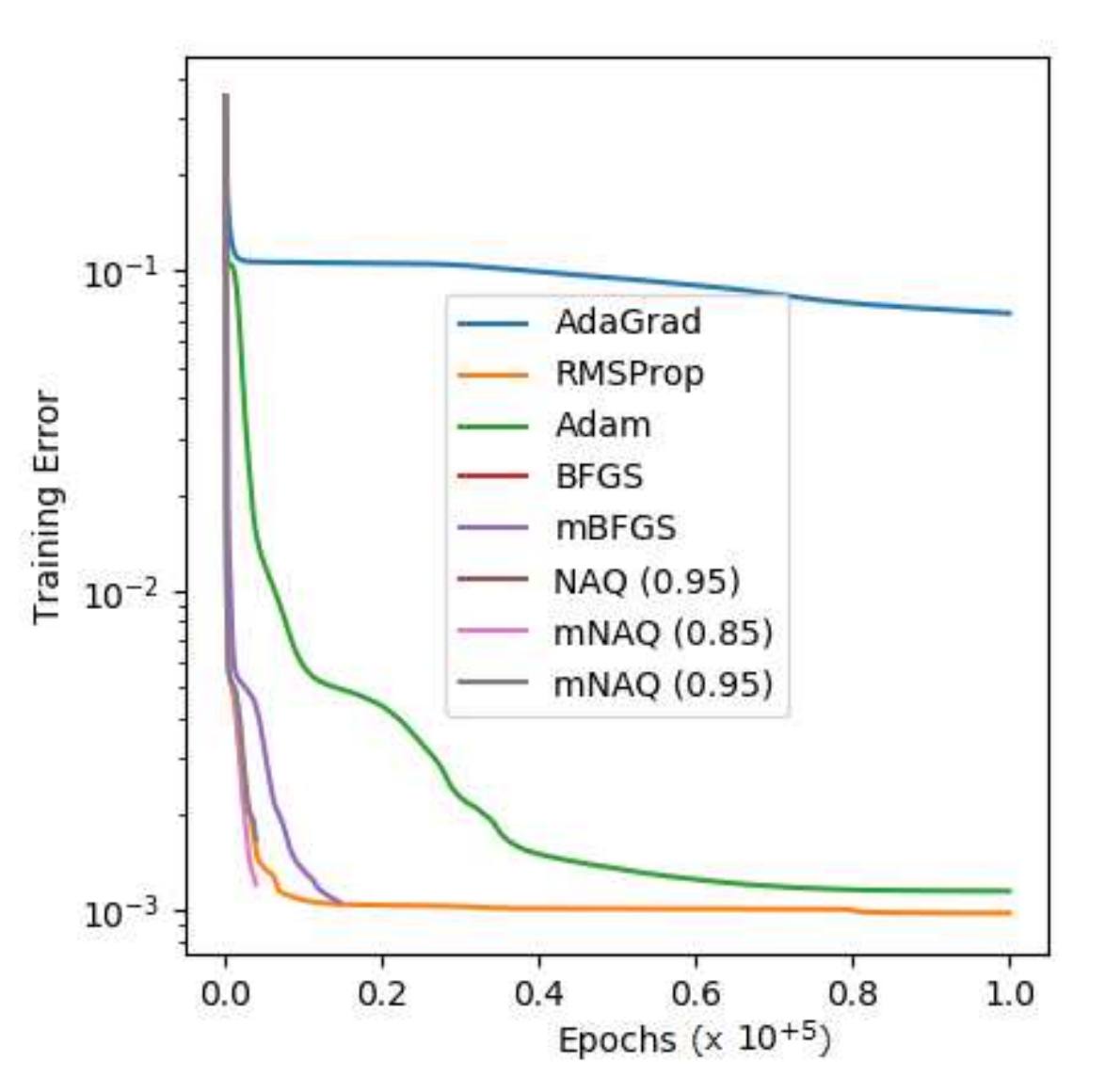}
\end{center}
\vspace{-0.5cm}
\caption{The average training errors $vs$ iteration count for WGF.}
\label{fig:scanner}

\end{figure}

\subsection*{\indent$<{\it Example~2: Modelling~of~microstrip~ lowpass~filter}>$}

Further we evaluate the performance of the proposed algorithm on a larger microwave circuit problem to model a microstrip low pass filter (LPF). The dielectric constant and height of the substrate of the LPF are 9.3 and 1mm, respectively. Fig. 7 shows the layout of the microstrip LPF. The inputs to the neural network are the length D and frequency $f$. The outputs are the magnitudes of the S-parameters $|S_{11}|$ and $|S_{21}|$.The frequency range is 0.1 to 4.5 GHz. For the training and test data, length $D$ ranges between 12-20 mm and 13-19 mm respectively at intervals of 2mm. Each interval contains 221 samples. The training set comprises of 1105 samples and test set comprises 884 samples. The training and test data were generated using Sonnet [17]. The number of hidden neurons used is 45. Fig. 8 shows the training data of the microstrip low pass filter. Table IV shows the summary of simulation results. From the table, we observe that the second order methods result in lower training errors compared to the first order algorithms. Fig. 9 shows the average training error over epochs. Though the training errors of mBFGS and mNAQ are comparable, mNAQ converges much faster compared to mBFGS. mNAQ with $\mu= 0.8$ performs the best. Fig. 10 illustrates the output of the trained neural network with mNAQ $\mu = 0.8$ for two sets of lengths $d = 13$mm and 15mm. The output of the trained model is close to the original test dataset. Thus, we can conclude that the proposed algorithm can be used effectively in practical models.

\begin{figure}[htb]

\begin{center}
\includegraphics[width=4cm]{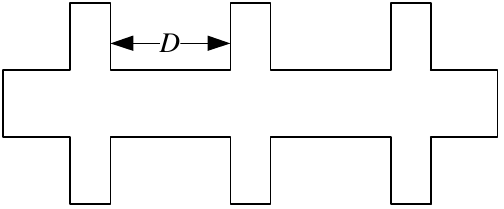}
\end{center}
\vspace{-0.3cm}
\caption{Layout of microstrip lowpass filter (LPF).}
\label{fig:scanner}

\end{figure}

\begin{figure}[htb]

\begin{center}
\vspace{-0.5cm}
\includegraphics[width=9cm]{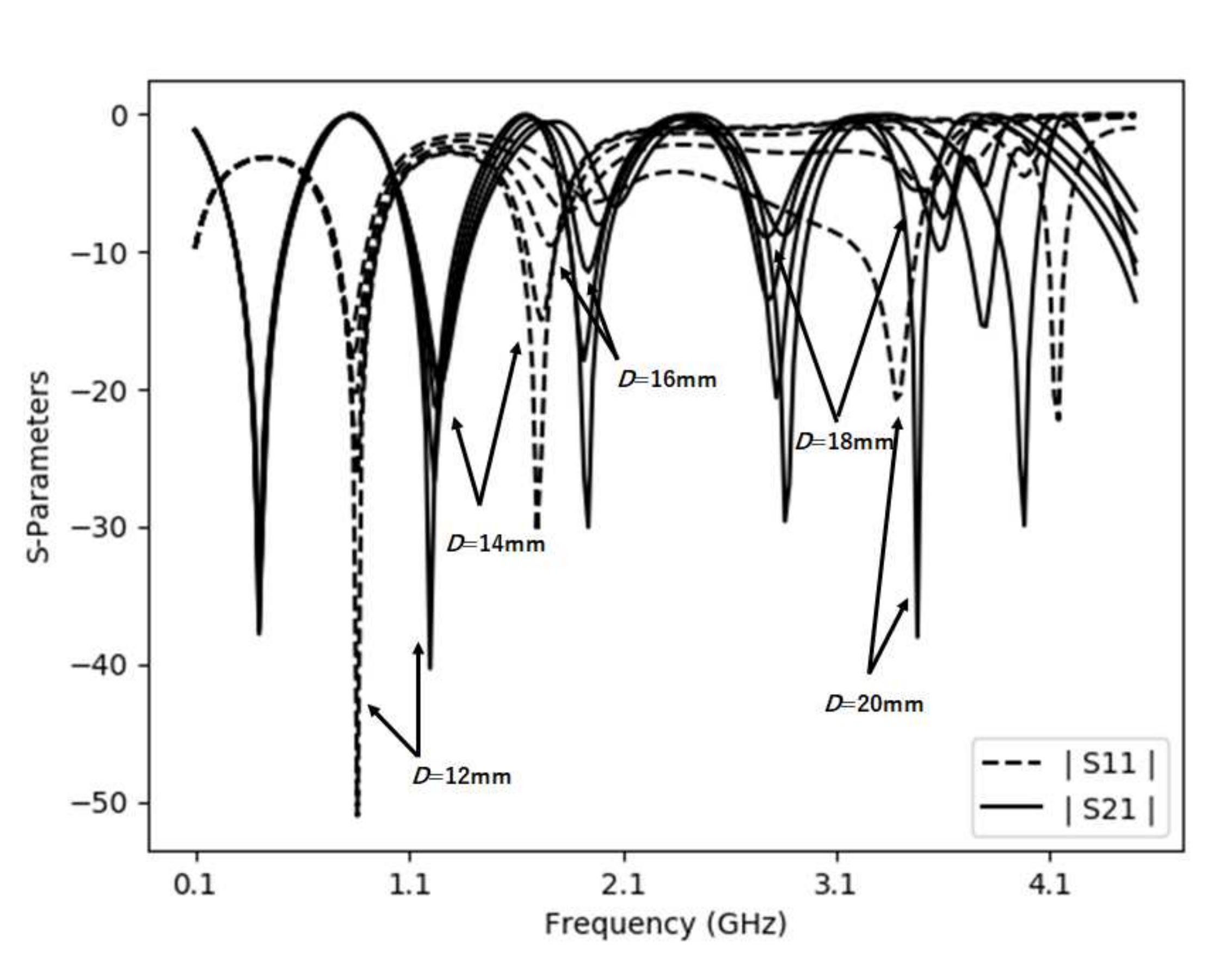}
\end{center}
\vspace{-0.5cm}
\caption{Training data set of microstrip lowpass filter (LPF).}
\label{fig:scanner}
\vspace{-0.5cm}
\end{figure}

\begin{table*}[htb]
\begin{center}
\caption{Summary of simulation results of microstrip low-pass filter (LPF).}
\label{tab:style}
\fontsize{9}{11}\selectfont
\begin{tabular}{ccccccc}
\hline 
Algorithm & $\mu$ & $E({\bf w})$($\times 10^{-3})$&Time & Iteration&$E_{test}({\bf w})$($\times 10^{-3})$  \\
&&Ave/Best/Worst& (s)& count&Ave/Best/Worst\\
\hline
AdaGrad & - & 26.6 / 26.4 / 26.7 & 112 & 100,000 & 22.4 / 22.3 / 22.5 \\
\hline
RMSprop & - & 2.99 / 2.44 / 4.07 & 113 & 100,000& 7.00 / 1.88 / 36.0  \\
\hline
Adam & - & 4.63 / 3.67 / 5.60 & 137 & 100,000& 37.0 / 3.41 / 212.5  \\
\hline
mBFGS & - &1.04 / 0.834 / 1.46 &493& 81,457& 1.01 / 0.529 / 3.52  \\
\hline
& 0.8 &0.93 / 0.827 / 1.37 &303 & 38,470& 0.744 / 0.534 / 1.07 \\
\cline{2-6}
mNAQ & 0.85 & 1.02 / 0.756 / 1.62 & 314 & 39,678 & 7.32 / 5.75 / 87.8 \\
\cline{2-6}
 & 0.9 &1.00 / 0.716 / 1.46 &242 & 30,619& 0.842 / 0.558 / 1.87  \\
\cline{2-6}
 & 0.95 & 1.24 / 0.834 / 1.85 &209 & 26,547& 2.08 / 0.600 / 13.7 \\
\hline
\end{tabular}
\end{center}
\end{table*}

\begin{figure}[htb]
\begin{center}
\includegraphics[width=7cm]{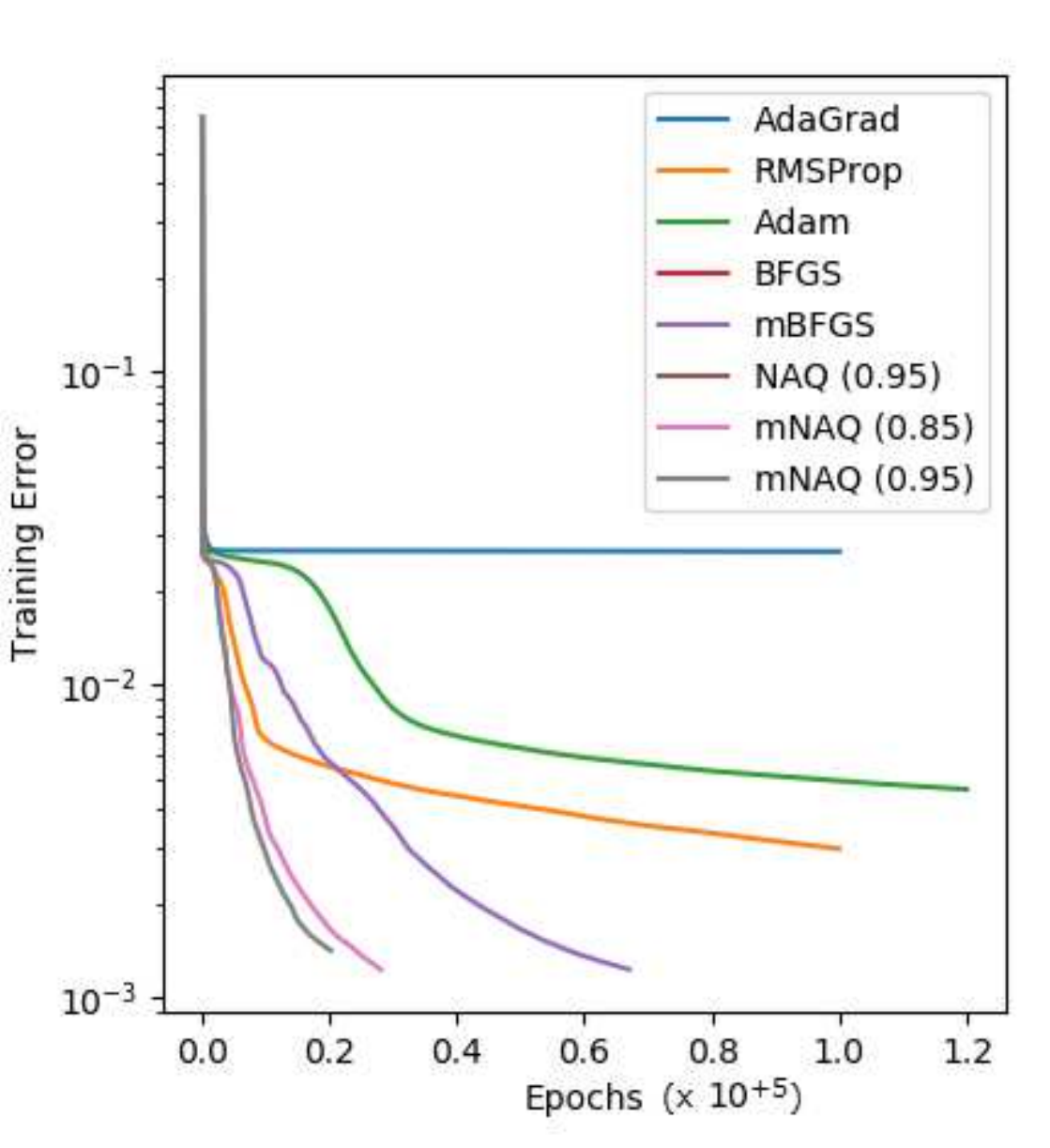}
\end{center}
\vspace{-0.5cm}
\caption{The average training errors $vs$ iteration count for LPF.}
\label{fig:scanner}

\end{figure}

\begin{figure}[htb]
\begin{center}
\includegraphics[width=8.5cm]{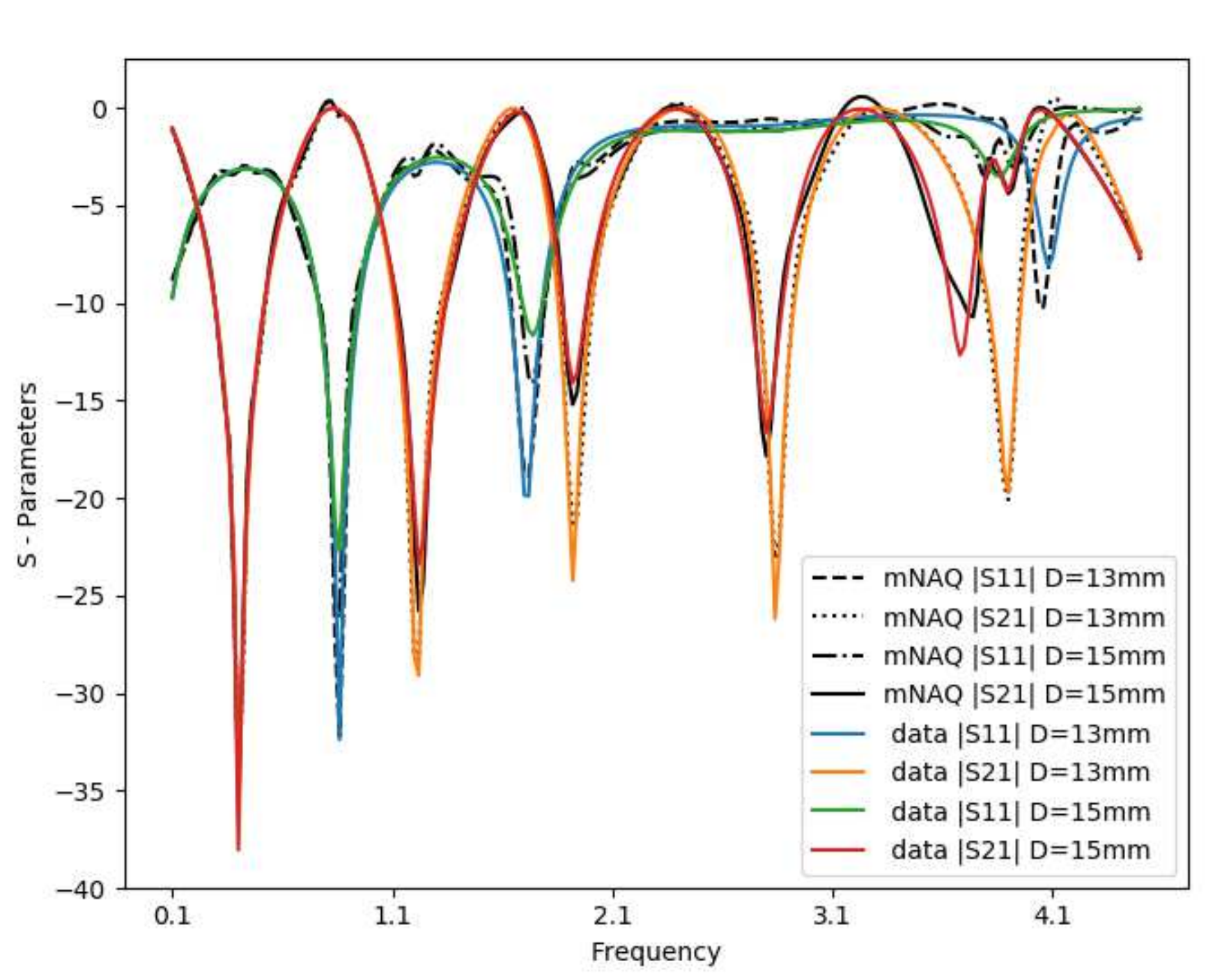}
\end{center}
\vspace{-0.5cm}
\caption{The comparison of network models of mNAQ $vs$ original test data of LPF.}
\label{fig:scanner}
\end{figure}

\section{Conclusions}

In this paper, we focus on implementing the BFGS, mBFGS and NAQ methods as a library in the Tensorflow environment. However, from the results obtained above, we observe that BFGS and NAQ does not terminate regularly while mBFGS and mNAQ performed normally. The line search satisfying Armijo's condition fails to find the step size after a few epochs, thus confirming that linesearch does not necessarily lead to global convergence.  However, this does not imply that the BFGS and NAQ methods with Armijo linesearch fails to converge. Upon adjusting the parameters appropriately, the BFGS and NAQ also converges to a stationary point. The proposed modified NAQ algorithm ensures global convergence without linesearch and can be effectively used in practical applications. The trained neural networks can be used as models of microwave devices in place of CPU-intensive EM/physics models to significantly speed up circuit design while maintaining good accuracies. Further with the distributed capabilities of TensorFlow to support both large-scale training and inference, we can conclude that we can effectively model complex problems and obtain much faster solutions.

\section*{Acknowledgment}
The authors thank Prof. Q.J. Zhang at Carleton University, Canada, for his support of microwave circuit models.

\end{document}